\documentclass[english]{IEEEtran}
\usepackage[T1]{fontenc}
\usepackage[latin9]{inputenc}
\usepackage{babel}
\usepackage{url}
\usepackage{amsmath}
\usepackage{amsthm}
\usepackage{amssymb}
\usepackage{graphicx}
\usepackage[unicode=true]
 {hyperref}
\usepackage{breakurl}

\makeatletter
\theoremstyle{plain}
\newtheorem{thm}{\protect\theoremname}
\theoremstyle{plain}
\newtheorem{lem}[thm]{\protect\lemmaname}

\usepackage{graphicx}

\makeatother

\providecommand{\lemmaname}{Lemma}
\providecommand{\theoremname}{Theorem}

\begin{document}

\title{Random Projection and Its Applications}

\author{\IEEEauthorblockN{Mahmoud Nabil}\\\IEEEauthorblockA{Department of Electrical and Computer Engineering, Tennessee Tech.
University, TN, USA 38505}\\\IEEEauthorblockA{\href{mailto:mnmahmoud42@students.tntech.edu}{mnmahmoud42@students.tntech.edu}}}
\maketitle
\begin{abstract}
Random Projection is a foundational research topic that connects a
bunch of machine learning algorithms under a similar mathematical
basis. It is used to reduce the dimensionality of the dataset by projecting
the data points efficiently to a smaller dimensions while preserving
the original relative distance between the data points. In this paper,
we are intended to explain random projection method, by explaining
its mathematical background and foundation, the applications that
are currently adopting it, and an overview on its current research
perspective.

\end{abstract}

\begin{IEEEkeywords}
Big Data, Random Projections, Dimensionality Reduction
\end{IEEEkeywords}

\section{Introduction}

Data transformation and projection is fundamental tool that is used
in many application to analyze data sets and characterize its main
features. Principal component analysis (PCA) for square matrices,
and its generalization Singular-value decomposition (SVD) for rectangular
real or complex matrices are examples of orthogonal data transformation
techniques that are used in many fields such as signal processing
and statistics. They are used to transform sparse matrices to condensed
matrices in order to get high information density, pattern discovery,
space efficiency and ability to visualize the data set. Despite their
popularity, Classical dimensionality reduction tecniques have some
limitations. First, the resultant directions of projection are data
dependent which make problems when the size of the data set increased
in the future. Second, they require high computational resources as
so it is impractical for high dimensional data. For instance, R-SVD
one of the fastest algorithms for SVD requires $O(km^{2}n+k'n^{3})$
\cite{golub2012matrix} for $m\times n$ matrix (where $k$ and $k'$
are constants). Third, in some applications access to the data is
restricted to streams where only frame sequences are available every
period of time. Last, these algorithms approximate data in low dimensional
space but not near a linear subspace. 

Random projection were presented to address these limitation where
the idea is to project data points to random directions that are independent
on the dataset. Random projection is simpler and computationally faster
than classical methods especially when the dimensions increased. Regarding
the computational requirement for random projection, it is $O(dmn)$
for $m\times n$ matrix \cite{bingham2001random}, where $d$ is the
size of the projected dimensions. This means that it compromises between
the processing time and a controlled accuracy for the intended application.
An interesting fact about random projection is that it can preserve
distance between the original and the projected data points with high
probability. And therefore, beside the geometric intuition for random
projection, it can be viewed as a local sensitivity hashing method
that can be used for data hiding and security applications \cite{liu2006random,jassim2009improving,yang2010dynamic}. 

\begin{figure}
\center \includegraphics[scale=0.25]{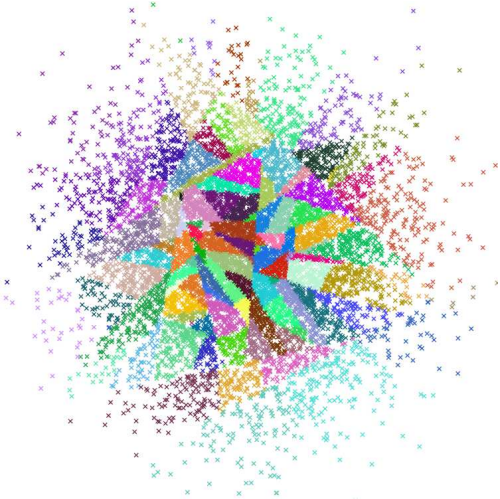}

\caption{Random projection using approximate nearest neighbors method \label{fig:Random-projection-using}}

\end{figure}

Another task which frequently involves random projection when the
data dimensionality is high, is nearest neighbor search where the
target is to return a group of data points which are closely related
to a given query. One can argue here why textual search methods like
inverted index can work on large document data sets but can't work
for images. This for two main reasons. First, textual data are sparse
which means if you picked up any document it only contains a few set
of tokens from the language vocabulary, however for images, data are
dense where for any image the useful pixels spans most of the image.
Second, the tokens themselves are the features to the document, where
only two or three words are enough to describe the document unlike
the pixels. These reasons make random projection more appealing for
nearest neighbor searching applications. The idea is that, For a given
search query instead of doing a similarity matching brute force search
for all data points in our dataset, we are only need to search in
the region that surrounds our query. The searching is done in two
stages namely: candidate selection, and candidates evaluation where
every data point in the new search space are evaluated. The core idea
is to partition the search space into dynamic variable size regions.
This force close data points to be mapped to the same regions which
increases their probabilities to be as candidates for a given search
query in the same region. In addition, to further increase the search
success rate, the search region can be partitioned several time depending
on the required accuracy and the processing time. Figure \ref{fig:Random-projection-using}
shows an example of random projection using approximate nearest neighbors
method on two dimensional data where regions have different colors.

In practice, some companies have utilized random projection into their
systems. Spotify a digital music platform \footnote{\url{https://www.spotify.com/}}
uses this method to find the approximate nearest neighbors music recommendations
\cite{arya1994optimal} as a part of their open source system \footnote{\url{https://github.com/spotify/annoy}}
. Esty is an E-commerce platform \footnote{\url{https://www.etsy.com/}}
uses random projection for user/product recommendation, their model
can be adapted in other ways, such as finding people with similar
interests, finding products that can be bought together and so on. 

Random projection is based upon the Johnson-Lindenstrauss lemma \cite{johnson1984extensions}
proposed in 1984 which states that \emph{``A set of points in a high-dimensional
space can be projected into a lower dimension subspace of in such
a way that relative distances between data points are nearly preserved''.
}It should be noted that the lower dimension subspace is selected
randomly based on some distribution. Furthermore, some recent and
faster algorithms rely on this lemma will be also discussed in this
paper.\emph{ }

The remainder of this paper is organized as follows. The mathematical
background and theorem proof are discussed in Section \ref{sec:Mathematical-Background}.
Some faster and computationally efficient random projections methods
are discussed in Section \ref{sec:Computationally-Efficient-Method}.
Applications and the current research perspective are discussed in
Section \ref{sec:Applications-and-Current}. Finally we draws our
conclusion in Section \ref{sec:Conclusion}.

\section{Mathematical Background \label{sec:Mathematical-Background}}

The ultimate aim of any data transformation on any data transformation/projection
technique is to perserve as much information as possible between the
original and the transformed data sets while better presents the data
in its new form. An essential step towards the proof of the random
projection of a a vector $v\in\mathbb{R}^{d}$where $d$ is typically
large to a k-dimensional space $\mathbb{R}^{k}$ is the Johnson\textendash Lindenstrauss
lemma \cite{johnson1984extensions} below:
\begin{lem}
For any small value $0<\epsilon<1$ and a set of $V$ of $n$ points
in $\mathbb{R}^{d}$, $\exists f:\mathbb{R}^{d}\rightarrow\mathbb{R}^{k}$
such that $\forall u,v\in V$ the following inequality holds with
high probability:

\[
(1-\epsilon)\sqrt{k}|v_{i}-v_{j}|\leq|f(v_{i})-f(v_{j})|\leq(1+\epsilon)\sqrt{k}|v_{i}-v_{j}|
\]
\end{lem}
The previous lemma act as a limiting bound (sandwich) for the distance
between the projected vectors $|f(v_{i})-f(v_{j})|^{2}$ and the distance
of the original vectors $|v_{i}-v_{j}|^{2}$.

\textbf{Proof. }Assume with out loss of generality the projection
function $f:\mathbb{R}^{d}\rightarrow\mathbb{R}^{k}$ is given by
\[
f(v)=(u_{1}.v,u_{2}.v\dots,u_{k}.v)
\]
where each $u_{i}\in\mathbb{R}^{d}$ is a Gaussian vector. In addition
assume that $|v|=1$.

Step 1. Each $u_{i}.v$ value is an independent Gaussian random variable
with zero mean and unit variance. This can be easily proved, since
$u_{i}.v=\sum_{j=1}^{d}u_{ij}v_{j}$ is sum independent Gaussian random
variables, therefore, the result random variable $u_{i}.v$ is also
Gaussian with mean equal the sum the individual means, and variance
can be obtained as the following

\[
Var(u_{i}\text{·}v)=Var(\sum_{j=1}^{d}u_{ij}v_{j})=\sum_{j=1}^{d}v_{j}^{2}Var(u_{ij})=|v|=1
\]

Step 2. According to Gaussian Annulus Theorem \cite{blum2016foundations}
for any high dimension Gaussian vector $x\in\mathbb{R}^{d}$, and
for $\beta\leq\sqrt{d}$, $1-3e^{-c\beta^{2}}$of the probability
mass lies within the annulus $\sqrt{d}-\beta\leq|x|\leq\sqrt{d}+\beta$
this can be written as
\begin{equation}
Prob(||x|-\sqrt{d}|\geq\beta)\leq3e^{-c\beta^{2}}\label{eq:Gaussian Annulus Theorem}
\end{equation}

Applying Gaussian Annulus Theorem \ref{eq:Gaussian Annulus Theorem}
to the Gaussian vector $f(v)\in\mathbb{R}^{k}$ and setting $\beta$
to $\epsilon$$\sqrt{k}\leq\sqrt{k}$ we got 

\[
Prob(|f(v)|-\sqrt{k}\geq\epsilon\sqrt{k})\leq3e^{-c\epsilon^{2}k}
\]

Multiplying inner inequality by $|v|=1$

\begin{equation}
Prob(||f(v)|-\sqrt{k}v|\geq\epsilon\sqrt{k}v)\leq3e^{-c\epsilon^{2}k}\label{eq:Random Projection Theorem}
\end{equation}

The latter equation is called the random projection theorem, which
bounds the upper bound of the probability that the difference between
the projected vector and the original vector shall exceeds a certain
threshold. What interesting is that with high probability $|f(v)|\approx\sqrt{k}|v|$.
So to estimate the difference between any two projected vectors $v_{1}$
and $v_{2}$ , we can calculate 
\[
f(v_{1}-v_{2})=\frac{f(v_{1})-f(v_{2})}{\sqrt{k}}\approx v_{1}-v_{2}
\]

Step3. By Applying the Random Projection Theorem \ref{eq:Random Projection Theorem}
the difference that $|f(v_{i})-f(v_{j})|$ is bounded by the range

\[
[(1-\epsilon)\sqrt{k}|v_{i}-v_{j}|,(1+\epsilon)\sqrt{k}|v_{i}-v_{j}|]
\]

with probability 
\[
1-3e^{-c\epsilon^{2}k}
\]

Two interesting facts we have from this poof. First, the number of
projected dimensions $k$ is completely independent on the original
number of dimension $d$ in the space, and it can be proved that it
only depends on the number of points in the dataset in a logarithmic
form and the selected error threshold $\epsilon$ where 
\begin{equation}
k\text{\ensuremath{\geq}}\frac{3\ln n}{c\epsilon^{2}}\label{eq:dimension bound}
\end{equation}

However, the error $\epsilon$ has a quadratic effect in the denominator
of equation \ref{eq:dimension bound} which means for $0.01$ error,
$k$ should be in the range of tens of thousands which is very high.
Second, unlike PCA and SVD the projection function is independent
on the original data completely. In addition, the $k$ projection
dimensions don't need to be orthogonal.

\section{Computationally Efficient Methods \label{sec:Computationally-Efficient-Method}}

Despite the simplicity of the random projection method as we showed
in section \ref{sec:Mathematical-Background}, in some applications
such as databases the proposed method may be costly. So Achlioptas
\cite{achlioptas2001database} proposed a new method that is computationally
efficient for this kind of applications. Achlioptas show that for
a random $d\times k$ transformation matrix $T$, where each entry
$t_{ij}$ of the matrix is independent random variable that follow
one of the following very simple probability distributions

\[
t_{ij}=\begin{cases}
+1\begin{aligned}\quad\end{aligned}
\begin{aligned}with\:probability\end{aligned}
 & 0.5\\
-1\begin{aligned}\begin{aligned}\quad\end{aligned}
\end{aligned}
\begin{aligned}\dots\end{aligned}
 & 0.5
\end{cases}
\]

\[
t_{ij}=\sqrt{3}\times\begin{cases}
+1\begin{aligned}\quad\end{aligned}
\begin{aligned}with\:probability\end{aligned}
 & 1/6\\
\begin{aligned}\end{aligned}
0\begin{aligned}\quad\end{aligned}
\begin{aligned}\quad\end{aligned}
\begin{aligned}\dots\end{aligned}
 & 2/3\\
-1\begin{aligned}\begin{aligned}\quad\end{aligned}
\end{aligned}
\begin{aligned}\dots\end{aligned}
 & 1/6
\end{cases}
\]
with probability at least $1-n^{-\beta}$ and for all vectors in the
database Johnson-Lindenstrauss lemma is satisfied. This method is
very efficient due to the use of the integer arithmetic in the calculations. 

\section{Related Applications and Current Research \label{sec:Applications-and-Current}}

Sparse recovery is an inverse problem to random projection and it
is the basic building block behind compressed sensing and matrix completion.
In this section we define each of the application and by showing how
they were inspired by the random projection idea.

\subsection{Compressed Sensing}

\begin{figure}
\center \includegraphics[scale=0.5]{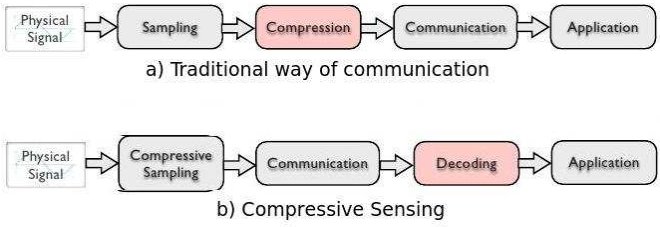}

\caption{Traditional communication architecture versus compressed sensing architecture.
\label{fig:Traditional-communication-archit}}
\end{figure}

According to Shannon-Nyquist sampling theorem, in order to be able
to reconstruct a signal with bandwidth $B$ from its samples, we need
a sampling rate $2B$. In compressed sensing, a very low sampling
rate can be used while the signal construction is achievable.

Let's consider a camera with 10 Megabyte pixels resolution that capture
a high quality image then it automatically converts it to a storage
efficient extension such as JPEG so that the resultant image can be
stored in a compressed format of about 100 Kilobyte with about the
same acceptable human eye resolution. This seems as a large waste
of the captured data. The idea is that, unlike the traditional way
of acquiring a high quality measurements then store them in an efficient
way, compressed sensing is working in a different way as shown figure
\ref{fig:Traditional-communication-archit} the sampling and the compression
stages are merged together and the receiver has to decode the incoming
message. In compressed sensing, each sensor acquire a very low quality
measurement for example a '\emph{Single-pixel Camera}' \cite{duarte2008single},
nevertheless, we should be able to combine and decompress all the
sensed data and get an acceptable quality compared to the 10 Megabyte
camera. In nutshell, the classical overview of sensing was to measure
as much data as possible, which is very wasteful. In compressed sensing,
the idea is to take $m$ random measurements then with high probability
we are still able to reconstruct the measured signal. In \cite{candes2006robust}
Candes and Tao proposed the Exact Reconstruction Principle, that gives
a new bounds for reconstructing any signal using its random compressed
samples.

Lets consider a discrete time signal $f\in\mathbb{R}^{n}$. In addition,
assume $\Psi\in\mathbb{R}^{n\times n}$ be a basis matrix where $\psi_{i}\in\mathbb{R}^{n}$.
So any signal y can be represented as a linear combination of the
columns of $\Psi$. In particular, suppose that our signal is defined
by
\begin{align*}
f & =\sum_{i=1}^{n}\psi_{i}x_{i}\\
 & =\Psi x
\end{align*}

where $x\in\mathbb{R}^{n}$ is a sparse coefficient vector to determine
the significant of the basis vector $\psi_{i}$. 

We can measure $f$ by taking few random measurements 
\begin{equation}
y_{j}=\phi_{j}^{T}f=\phi_{j}^{T}\Psi x\label{eq:signal}
\end{equation}

where $\phi_{j}\in\mathbb{R}^{n}$ is the $jth$ compressed sensing
vector $1<j<m$. We can deduce that if the noise is zero, then at
least $n$ measurements vectors $\phi_{j}$ are needed to be able
to reconstruct the signal $f$. 

Using compressed sensing we are able to get a tighter bound on the
number of measurements $y_{j}$ that we should have to reconstruct
$f$. This bound is $O(S\log(n/S))$, where $S$ is the number of
non-zero elements in the vector $x$. 

If we can use only $m<<n$ measurements using a measurement matrix
$\Phi\in\mathbb{R}^{m\times n}$. Then, equation \ref{eq:signal}
can be written in matrix form as

\[
y=\Phi f=\Phi\Psi x=Ax
\]

where $y\in\mathbb{R}^{m}$ is the measurements vector and let $A=\Phi\Psi$.
Using \emph{restricted isometry property} (RIP) \cite{candes2005decoding},
we can define isometry constant $\delta_{s}$ such that

\begin{equation}
(1-\delta_{s})|x|\leq y=|Ax|\leq(1+\delta_{s})|x|\label{eq:RIP}
\end{equation}

The Johnson-Lindenstrauss embedding property implies the Restricted
Isometry Property (RIP) above. We say that the matrix $A$ have RIP
of order $S$. However, if $A$ has order $2S$, we can measure two
compressed vectors $y^{(1)}$and $y^{(2)}$ $\in\mathbb{R}^{m}$ we
can easily get the following inequality 

\begin{align}
(1-\delta_{s})|x^{(1)}-x^{(2)}| & <|A(x^{(1)}-x^{(2)})|<(1+\delta_{s})|x^{(1)}-x^{(2)}|\nonumber \\
(1-\delta_{s})|x^{(1)}-x^{(2)}| & <|y^{(1)}-y^{(2)}|<(1+\delta_{s})|x^{(1)}-x^{(2)}|\label{eq:RIP2}
\end{align}

where $x^{(1)}-x^{(2)}$ is at most $2S$ sparse vector. Hence, if
we can enumerate all the $2S$ sparse vectors and compare each of
them to $|y^{(1)}-y^{(2)}|$, the original signal can be easily reconstructed.
We can see the analogy between equation \ref{eq:RIP} and Random Projection
Theorem. It is like a linear algebra problem if you solve it correctly,
the original signal can be reconstructed. However, due to the random
noise the reconstruction is more difficult problem \cite{lukas2006digital,metzler2016denoising}.
It is also worth to mention that, one of the foundation of compressed
sensing research was to prove that the randomly generated sensing
matrix $\Phi$ follow the RIP criteria. In \cite{baraniuk2008simple}
Baraniuk et al. aimed to give a condition for different random sensing
matrices to follow RIP criteria. In addition, it was proved that a
random matrix that follow a Gaussian distribution, inherently obey
the RIP criteria. 

\subsection{Matrix Completion}

\begin{figure}
\center \includegraphics[scale=0.25]{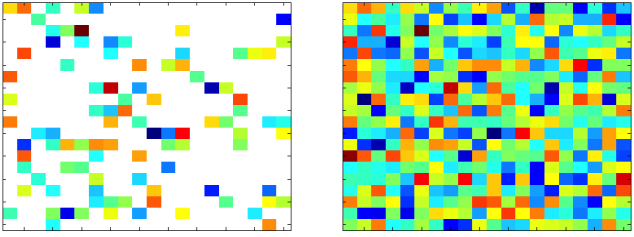}

\caption{Matrix completion example \label{fig:Matrix-completion-example}}
\end{figure}

Another interested task is the low rank matrix completion. It is used
in many applications like image in-painting where the goal is to recover
deteriorated pixels in an image as shown in figure \ref{fig:Matrix-completion-example}.
In addition, \emph{Netflix problem} where the goal is to complete
the customer-movie rating matrix given only the some customers rating,
in order to build a robust recommendation system. The \emph{Netflix
}one million dollar grand prize was given to \cite{koren2009bellkor}
BellKor team for their 10.06\% recommendation system.

Lets consider a partially observed matrix $Y\in\mathbb{R}^{m\times n}$,
we define the matrix completion problem as to find the minimum rank
matrix $X\in\mathbb{R}^{m\times n}$ that best approximates the matrix
$Y$. Removing this limitation, matrix completion problem has undetermined
solution because the missing values can be assigned any random values.
The mathematical formulation of the problem is defined by

\begin{align*}
min_{X\in\mathbb{R}^{m\times n}} & \quad rank(X)\\
s.t.\quad & X_{ij}=Y_{ij}\quad for\:observed\:locations\:(i,j)
\end{align*}

In general the rank minimization is NP-hard problem. However, in \cite{candes2009exact}
Candes et al. proposed a convex relaxation solution to the problem
to minimize the nuclear norm $||X||_{*}$ which is defined as the
sum of the singular values of $X$. Candes proposed some assumptions
on the number of the observed entries in $Y$ so that $X$ can be
recovered with high probability. The nuclear norm minimization is
given by

\begin{align}
min_{X\in\mathbb{R}^{m\times n}} & \quad||X||_{*}=\sum_{i=1}^{m}\sigma_{i}(X)\nonumber \\
s.t.\quad & X_{ij}=Y_{ij}\quad for\:observed\:locations\:(i,j)\label{eq:Matrix Completion}
\end{align}

The assumptions that are proposed to solve the matrix completion problem
are:
\begin{enumerate}
\item The observed entries are uniformly sampled from all subsets of entries.
\item Coherence: where the goal is to try to align the rows and/or the columns
of $X$ with the basis vectors. We are interested in low coherence
subspace, where if we assumed column and row spaces are $U$ and $V$
then $max(\mu(U),\mu(V))\le\mu0$ for some positive value $\mu_{0}$
where $\mu$ is the coherence factor. In addition, The matrix $\sum_{1\leq k\leq r}u_{ik}v_{jk}$
should have an upper bound on its entries by $\mu_{1}\sqrt{r/(n_{1}n_{2})}$
where $n_{1}$ and $n_{2}$ are the matrix dimensions.
\item Number of observed entries: this sets a lower bound on the number
of the observed elements $m$ in $X$ so that the completion is possible.
In \cite{candes2009exact}, Candes proved that this lower bound is
\[
m\geq Cmax(\mu_{1}^{2},\mu_{0}^{1/2}\mu_{1},\mu_{0}n^{1/4})nr(\beta\log n)
\]
 where $C$ and $\beta$ are constants and $\mu_{1}=\mu_{0}\sqrt{r}$.
\end{enumerate}
For $\beta>2$ equation \ref{eq:Matrix Completion} is solvable and
it is equal to $Y$ with high probability $1-cn^{-B}$.

\subsection{Human Activity Recognition}

Tracking the state and the actions of elderly and disabled people
using some sensors attached to their bodies has considerable importance
in health-care applications. It can facilitate the monitoring and
the detecting of any abnormal condition at the patient body and report
it. In \cite{damavsevivcius2016human} the authors proposed a method
that is working offline and it can recognize of daily human activities.
The system has three main stages: (a) de-noising sensor data (b) feature
extraction and feature dimensionality reduction using computationally
efficient random projection presented in section \ref{sec:Computationally-Efficient-Method}
(d) classification using Jaccard distance between kernel density probabilities.
The reported results on the USC-HAD dataset (Human Activity Dataset)
is within-person classification of 95.52\% and inter-person identification
accuracy of 94.75\%.

\subsection{Privacy Preserving Distributed Data Mining}

In many data mining applications such as health care, fraud detection,
customer segmentation, and bio-informative privacy and security concerns
have an immense importance due to dealing with different types sensitive
data. This call for privacy preserving techniques that can work on
encrypted or noisy data while being able to compute accurately and
efficiently a set of predefined operations such as Euclidean distance,
dot product, and correlation etc. In \cite{liu2006random} the authors
introduced data perturbation technique using random projection transformation
where some noise is added to the data before being sent to the cloud
server. The proposed technique preserves the statistical properties
of the dataset and also allows the dimensionality reduction of it.
It is considered as value distortion approach where the all data entries
are perturbed directly and at once (i.e. not independently) using
multiplicative random projection noise. The advantage of this technique
is that many elements are mapped to one element, which is totally
different from the traditional individual data perturbation technique,
and, therefore, it is even harder for the adversary to reconstruct
the plain text data. The technique depends on some lemmas explained
as follows
\begin{lem}
\emph{For random matrix $R\in\mathbb{R}^{p\times q}$ where all entries
$r_{i,j}$ are independent and identically chosen from gaussian distribution
with zero mean and $\sigma_{r}^{2}$ variance then}

\[
E(R^{T}R)=p\sigma_{r}^{2}I,\;and\:E(RR^{T})=q\sigma_{r}^{2}I
\]

\textbf{\emph{Proof. }}\emph{lets proof the first inequality. Assume
$\epsilon_{i,j}$ is the entry from $R^{T}R$ then}

\begin{align*}
\epsilon_{i,j} & =\sum_{t=1}^{p}r_{i,t}r_{t,j}\\
E(\epsilon_{i,j}) & =E(\sum_{t=1}^{p}r_{i,t}r_{t,j})\\
 & =\sum_{t=1}^{p}E(r_{i,t}r_{t,j})\\
 & =\sum_{t=1}^{p}E(r_{i,t})E(r_{t,j})\\
 & =\begin{cases}
\sum_{t=1}^{p}E(r_{i,t})E(r_{t,j}) & i\neq j\\
\sum_{t=1}^{p}E(r_{t,i}^{2}) & i=j
\end{cases}\\
 & =\begin{cases}
0 & i\neq j\\
p\sigma_{r}^{2} & i=j
\end{cases}
\end{align*}
\begin{lem}
\emph{for any two data sets $X\in\mathbb{R}^{m_{1}\times n}$ and
$Y\in\mathbb{R}^{m_{2}\times n}$, and let random matrix $R\in\mathbb{R}^{p\times q}$
where all entries $r_{i,j}$ are independent and identically chosen
from unknown distribution with zero mean and $\sigma_{r}^{2}$ variance,
also let $U=\frac{1}{\sqrt{k}\sigma_{r}}RX$,$V=\frac{1}{\sqrt{k}\sigma_{r}}RY$
, then}

\[
E(U^{T}V)=X^{T}Y
\]
\end{lem}
\emph{The} \emph{above results enables the following statistical measurements
(distance, angle, correlation) to be applied to the hidden data knowing
the original vectors are normalized }
\end{lem}
\begin{align*}
dist(x,y) & =\sqrt{\sum_{i}(x_{i}-y_{i})^{2}}\\
 & =\sqrt{\sum_{i}x_{i}^{2}+\sum_{i}y_{i}^{2}-2\sum_{i}x_{i}y_{i}}=\sqrt{2-2x^{T}y}
\end{align*}

\[
cos\theta=\frac{x^{T}y}{|x|.|y|}=x^{T}y
\]
\[
\rho_{x,y}=x^{T}y
\]
Thus, the number of attributes of the data can be reduced by random
projection and the statistical dependencies among the observations
will be maintained. It is worth to mention that, given only the projected
data U or V , original data can not be retrieved as the number of
possible solutions are infinite.

\textbf{For error analysis}, it can easily be proven that the mean
difference and the variance difference between the projected and the
original data are given as

\[
E(u^{T}v-x^{T}y)=0
\]
\[
Var(u^{T}v-x^{T}y)\leq\frac{2}{k}
\]
It can be seen that, the error goes down as k increases. This implies
that at high dimension space, the technique works better.

\textbf{For privacy analysis. }two types of attacks are considered
\begin{enumerate}
\item The adversary tries to retrieve the exact values of the projected
matrix X or Y, the authors proved that when $m\geq2k-1$, even if
matrix R is disclosed the original matrices can not be retrieved.
\item The adversary tries to estimates matrix X or Y, if the distribution
of R is known, if the adversary generates $\hat{R}$ according to
the known distribution then 
\[
\frac{1}{\sqrt{k}\hat{\sigma_{r}}}\hat{R^{T}}u=\frac{1}{\sqrt{k}\hat{\sigma_{r}}}\hat{R^{T}}\frac{1}{\sqrt{k}\sigma_{r}}Rx=\frac{1}{k\sigma_{r}}\hat{\epsilon}x
\]

the estimation of any data element from the vector x is given by

\[
\hat{x_{i}}=\frac{1}{k\hat{\sigma_{r}}\sigma_{r}}\sum_{t}\hat{\epsilon_{i,t}}x_{t}
\]

The expectation and the variance can be calculated as

\[
E(\hat{x_{i}})=0
\]
\[
Var(\hat{x_{i}})=\frac{1}{k}\sum_{t}x_{t}^{2}
\]

So the adversary can only get a null vector centered around the zero.
\end{enumerate}
The authors considered three applications on their paper all of them
relies on the dot product estimation namely: distance estimation,
k-mean clustering, and linear perceptron. As a result, the random
projection-based multiplicative perturbation technique keeps both
the statistical properties and the confidentiality of the data.

\section{Conclusion \label{sec:Conclusion}}

In this paper, we explained the random projection and the mathematical
foundation behind it. In addition, we explained some related applications
such as compressed sensing which made a breakthrough in the traditional
communication theorems where a very low sampling rate can be used
while the signal construction is achievable. Also, we explained the
matrix completion problem that is a basis for many data mining tasks
such as recommendation systems and image in-painting algorithms. 

\bibliographystyle{IEEEtran}
\bibliography{biliography}

\end{document}